%% file: conference_101719.tex
\documentclass[conference]{IEEEtran}
\IEEEoverridecommandlockouts
\usepackage{cite}
\usepackage{amsmath,amssymb,amsfonts}
\usepackage{algorithmic}
\usepackage{graphicx}
\usepackage{textcomp}
\usepackage{xcolor}
\usepackage{tabularx}
\usepackage{booktabs}

\def\BibTeX{{\rm B\kern-.05em{\sc i\kern-.025em b}\kern-.08em
    T\kern-.1667em\lower.7ex\hbox{E}\kern-.125emX}}
\usepackage{url}

\begin{document}

\input{title}
\input{abstract}
\input{introduction}
\input{related_work}
\input{methodology}
\input{experiments}
\input{results}
\input{conclusion}

\bibliographystyle{IEEEtran}
\bibliography{bibliography.bib}

\end{document}

%% file: title.tex
\title{WordAlchemy: A Transformer-based Reverse
Dictionary \\
\thanks{*These authors contributed equally to this work.}
}

\author{
\begin{tabular}[t]{c@{\extracolsep{8em}}c} 
 \textbf{Kanhaiya Balaji Madaswar\textsuperscript{*}}  & \textbf{Harshal Navneet Patil\textsuperscript{*}} \\
\textit{Dept. of Computer Engineering and }  & \textit{Dept. of Computer Engineering and }  \\
\textit{Information Technology} & \textit{Information Technology} \\
\textit{College of Engineering, Pune} & \textit{College of Engineering, Pune} \\
Maharashtra, India & Maharashtra, India 
\vspace*{1 cm}
\end{tabular}
\\
\begin{tabular}[t]{c@{\extracolsep{8em}}c} 
\textbf{Pranav Nitin Sadavarte\textsuperscript{*}}  & \textbf{Dr. Sunil B. Mane} \\
\textit{Dept. of Computer Engineering and}  & \textit{Dept. of Computer Engineering and}  \\
\textit{Information Technology} & \textit{Information Technology} \\
\textit{College of Engineering, Pune} & \textit{College of Engineering, Pune} \\
Maharashtra, India & Maharashtra, India 
\end{tabular}
}


\maketitle

%% file: abstract.tex
\begin{abstract}
A reverse dictionary takes a target word’s description as input and returns the words that fit the description. Reverse Dictionaries are useful for new language learners, anomia patients, and for solving common tip-of-the-tongue problems (lethologica). Currently, there does not exist any Reverse Dictionary provider with support for any Indian Language. We present a novel open-source cross-lingual reverse dictionary system with support for Indian languages. In this paper, we propose a transformer-based deep learning approach to tackle the limitations faced by the existing systems using the mT5 model. This architecture uses the Translation Language Modeling (TLM) technique, rather than the conventional BERT’s Masked Language Modeling (MLM) technique.

\end{abstract}


%% file: introduction.tex
\section{Introduction}

\begin{figure}[h]
\centering
\includegraphics[width=0.47\textwidth]{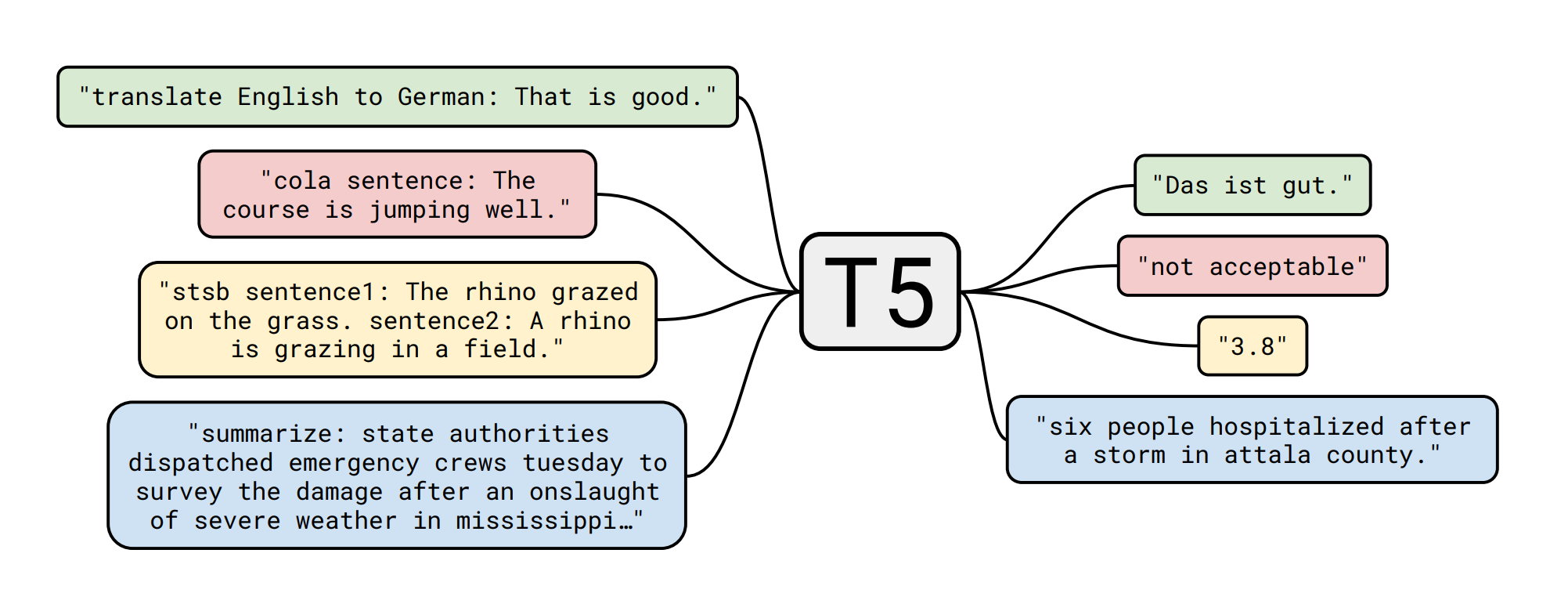}
\caption{T5: Text-to-Text Transfer Transformer \cite{conneau2019unsupervised}}
\end{figure}

Many times in our day-to-day life, we struggle to think of a correct word, even though we could describe it. Sometimes we can’t express our thoughts in single words even though we can describe the concept in phrases. This might be just a recalling problem or due to our lack of knowledge of a certain language. It is a very common problem usually faced when an individual tries to learn a new language. In such cases, dictionaries might not be the best solution. While dictionaries fulfill the needs of readers in terms of word definitions, they fall short in terms of addressing the needs of language producers (writers/speakers) in terms of finding a suitable word to match a thought in mind. 

This difficulty is handled by the ‘reverse dictionary’. It takes a user-provided description of the notion (in natural language) as input and outputs a set of terms that satisfy that definition. Although the problem is not new, it is of great concern due to the compromises that language creators must make, the most common of which is circumlocution (vagueness). Reverse Dictionaries have many practical usages e.g. Tip of the tongue (or lethologica) is a common phenomenon faced by people in which they fail to retrieve a word or term from memory, combined with partial recall and the feeling that retrieval is imminent. Sometimes, new language learners can describe a word in a particular language but fail to retrieve the exact word in the new language. Anomia patients, people who are able to recognize and describe an object but are not able to name it due to a neurological disorder, can also be assisted by using a reverse dictionary.

To address all these issues more accurately, we propose and develop a novel open-source Reverse Dictionary named “WordAlchemy”, mainly based on the proposed transformer-based mT5\cite{xue2021m} model.

%% file: related_work.tex
\section{Related Work}
In this section, we will focus on the previous work in the domain of Reverse Dictionary. We will study different algorithms and techniques used for experimenting and explore some previously built artificial intelligent models developed to solve the challenges in implementing Reverse Dictionary.

The most straightforward method to solve the problem is a database-driven approach. But this approach requires that the input phrase contain words that exactly match a dictionary definition, and it does not scale well. Online reverse dictionaries like Onelook and reversedictionary.org use algorithms based on this approach to provide results. But this approach will not be suitable for user descriptions and unseen data as it does not consider the semantic meaning of the description.

Many researchers tried to improve on the database-driven approach.
Shaw et al.(2013)\cite{shaw2013building} provided methods to overcome these challenges without affecting performance and output quality. They proposed a technique consisting of RMS (Reverse Mapping Set) generation. To further improve the accuracy of the results, it considers a set of conceptually similar words to generate RMS. It offers a significant improvement in performance and scale without affecting the results. Shete and Patil(2018)\cite{shete2018intelligent} enhanced it using an SVM model.
The SVM model was trained on WordNet Lexical Database, thus providing better conceptually accurate results. Thorat and Choudhari(2016)\cite{thorat2016implementing} presented a node-graph architecture. It generates a reverse map by generating a graph of conceptually similar words.

The researchers then moved on to creating a semantic approach, which will help better understand the meaning of the user input. M{\'e}ndez, Calvo, and Moreno-Armend{\'a}riz(2013)\cite{mendez2013a} proposed the creation of semantic space to represent words as vectors based on the semantic similarity measures and then used algebraic analysis to select a sample of candidate words.

Recent advancements in deep learning opened new possibilities in the domain of Reverse Dictionary. Hill et al.(2016)\cite{hill2016learning} first used RNN(Recurrent Neural Network) to solve this problem. It gave better results than the previously defined models. Morinaga and Kazunori(2018)\cite{morinaga2018improvement} improved on Hill's(2016)\cite{hill2016learning} model. It uses category inference to eliminate irrelevant words from the ranked result. Even though the category of the target word is undefined, it is inferred from the CNN Model. Park, Safwan, and Sharma(2018)\cite{park2018reverse} used LSTM(Long Short Term Memory) to capture contextual meaning in the input phrase and training dictionary.
Zhang et al.(2019)\cite{zhang2019multi} Adopted the bi-directional LSTM (BiLSTM) with attention as the basic framework and added four feature-specific characteristic predictors to it. The predictors include two internal (POS Tag and Morpheme predictor) and two external predictors (Word Category and Sememe). It views each characteristic predictor as an information channel for searching the target word.
Yan et al.(2020)\cite{yan2020bert} presented a transformer-based approach using the BERT architecture for monolingual and cross-lingual Reverse Dictionary. The architecture uses MLM(Masked Language Modeling) to predict the target word. They extended this method for cross-lingual Reverse Dictionary by translating the Chinese input text to English and then used the same model to predict the output. This method merely translates the words rather than considering the whole context of the sentence, which generates some translational error affecting the accuracy of the result.

%% file: methodology.tex
\section{Methodology}

The reverse dictionary task is to find the target word ‘w’ given its definition d = [w1, w2, . . . , wn], where, d and w should be in the same language.

\begin{figure}[ht]
\centering
\includegraphics[width=0.47\textwidth]{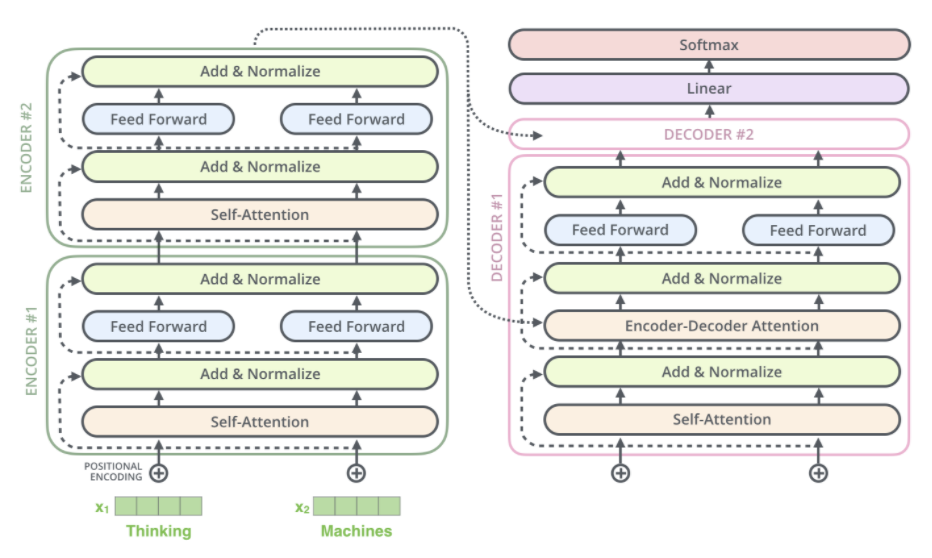}
\caption{T5: Encoder Decoder Model \cite{raffel2019exploring}}
\end{figure}

\subsection{T5}

T5\cite{raffel2019exploring} (Text-to-Text Transfer Transformer) is an encoder-decoder model that has been pre-trained on a multi-task mixture of unsupervised and supervised tasks. Being an encoder-decoder model, all the NLP tasks are reframed into a unified text-to-text format, where the input and output are always text strings.
For specific tasks, prefixes are prepended at the beginning of inputs and outputs. For example, “summarise” for text-summarisation, “translate” for language translation. These specific task name prefixes play an important role in identifying different tasks for the model.
For the pre-training objective, the paper defines three approaches. The first is language modeling, which just includes predicting a word by considering all the preceding words in the sentence. The second, is deshuffling, in which initially all the words of the sentence are shuffled, and then the model is instructed to predict the original text. Third, corrupting spans also known as the denoising objective, includes masking the sequences of words and training the model to predict those masked sequences. Using these pre-training objectives, T5 gives promising results after fine-tuning it.     

\subsection{mT5}

mT5 is a multilingual Transformer model that has been pre-trained on a dataset (mC4) that contains text in 101 languages. The mT5 model (based on T5) is developed to handle any Natural Language Processing task like classification, named-entity recognition, question answering, etc. by redefining the problem as a sequence-to-sequence process. 
The working of mT5 is similar to that of the T5 model. In simple terms, the text goes in and the text comes out. In a classification task, for example, the text sequence to be classified can be the model's input, and the model's output will be the sequence's class label. This is even more straightforward in terms of translation. The input text is in one language, and the output text is in a different language.

\subsection{T5 for monolingual reverse dictionary}

The main idea behind using T5 for the monolingual reverse dictionary is that it proposes reframing all NLP tasks as an input text to output text formulation. Each word instance in the dataset is made up of pair of words and definitions. Hence, while training the model we set the input as the definition of the word and the target set to the word itself. We define a ReverseDictionary class in which we build the model and make a call to tokenizers and data loaders.
Each instance is of input and the target is then prepended by their respective prefixes, “Definitions: ” and “Word: ”.
Each instance from the dataset is then converted into pairs of ids and masks. 
T5-base is used as the tokenizer, which is used to generate tokenized inputs and targets. For tokenization, we use BatchEncodePlus which returns us batch encoded PyTorch tensors padded to max\_length.
The T5FineTuner class includes the four basic functions of training a model, which is the data loaders, forward pass through the model, training one step, validation on one step as well as validation at epoch end functions.
While training at each forward step we generate the input\_ids (which is just the tokenized input text padded to max\_length), attention\_masks, labels, and decoder\_attention\_masks.
Next, the loss is calculated and AdamW is used as the optimizer. AdamW is an improvised version of the Adam optimizer in which only after regulating the parameter-wise step size is weight decay conducted. Finally, the model is validated against a validation dataset and final results are logged. 

\begin{figure}[ht]
\centering
\includegraphics[width=0.47\textwidth]{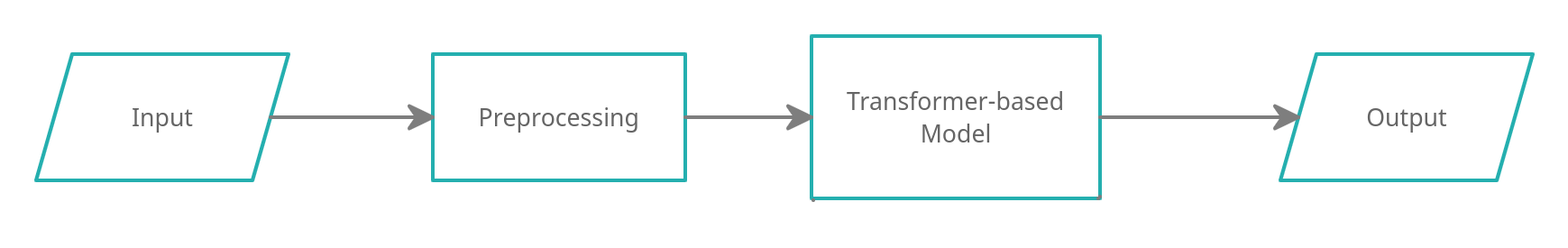}
\caption{Flowchart for ``Reverse Dictionary"}
\end{figure}

\subsection{mT5 for multilingual reverse dictionary}

For finetuning this model we use the same setup as that of the monolingual model with the only difference being the pre-trained model used. This setup uses the mT5 model, a multilingual version of the T5 model. T5's architecture and training technique are closely followed by mT5, although it is trained on mC4 (a whopping 26 Terabytes! ), a multilingual variant of the C4 dataset. It has all of the benefits of T5, but it also supports a remarkable 101 languages! The difference between this scenario and the monolingual one is that this setting uses a different pre-trained model i.e mt5-base. The token embedding of the mT5 model comprises subwords in several languages because it was trained in multiple languages. Unlike T5, mT5 needs to be fine-tuned before using it for any downstream tasks. As we have only one downstream task i.e. the reverse dictionary task, the prefix is of no use for fine-tuning.

%% file: experiments.tex
\input{query_table}

\section{Experimental Setup}

\subsection{Dataset}

For the monolingual T5 model, We use the English dictionary definition dataset created by Hill et al. (2016) \cite{hill2016learning} as the training set. The definitions are extracted from five electronic resources: WordNet \cite{fellbaum1998wordnet} \cite{miller1995wordnet}, The American Heritage Dictionary, The Collaborative
International Dictionary of English, Wiktionary and The dictionary by Merriam-Webster.
It contains about 100,000 words and 900,000 word-definition pairs. It is randomly divided into two different unique sets 1) Train set - which is used to finetune the pre-trained t5 model. 2) Test set - the finetuned t5 model is tested on this unseen data to find the performance of the created model. 
Hindi and Marathi datasets for the multi-lingual mT5 model are created from wordnet bindings for corresponding languages. Hindi and Marathi wordnets \cite{bhattacharyya2010indowordnet} \cite{chakrabarti2014an} are based on the similar idea of English WordNet. WordNet \cite{fellbaum1998wordnet} \cite{miller1995wordnet} is a Lexical Database made publicly available by Princeton University for research. WordNet \cite{fellbaum1998wordnet} \cite{miller1995wordnet} is more than a conventional dictionary. It is a system for bringing together different lexical and semantic relations between words. It organizes the lexical information in terms of word meanings and can be termed a lexicon based on psycholinguistic principles. Hindi and Marathi Wordnets \cite{bhattacharyya2010indowordnet} \cite{chakrabarti2014an} are released under GNU GPL 3.0 license. They are funded by T.D.I.L. (Technology Development for Indian Languages). Hindi Wordnet contains around 63,800 unique words, while Marathi WordNet contains around 23,000 unique words. Both these datasets are divided in a 9:1 ratio to form 1) a train set and 2) a test set, similar to the monolingual T5 model.

\subsection{Evaluation Metrics}

Referring to previous work (Hill et al., 2016 \cite{hill2016learning}), we have used 4 evaluation metrics for our model: the accuracy of the target word being present in the top 1/10/100 i.e Acc@1/10/100, finding the median of the rank of the target words (Lower the median, better the result), and the performance on human-written definitions. Each test is performed five times, and the average findings are reported.

\subsection{Hyper-parameter Settings}

The English T5 and Multilingual mT5 are derived from Raffel et al. (2020) and Kale et al. (2021). Both T5 and mT5 models are the base versions. For both the models, hyper-parameters are found by checking the performance on the development set and further evaluated on the test dataset. Some of the hyper-parameters we fine-tune are maximum sequence length, optimizer learning rate, weight decay regularization, adam epsilon, batch size, gradient accumulation steps, number of epochs, etc.

Both T5 and mT5 use the regular cross-entropy loss or log loss. In cross-entropy loss, we compare the probability of the predicted class and the actual class and finally generate an absolute score depicting the deviation of the predicted word from the actual word. Being a log function, small scores are penalized lowly and bigger scores are penalized greatly. This score is then used to adjust the weights of the model while training.
Once the input words are tokenized into input\_ids they are passed through the model and the decoder output labels are generated. The cross-entropy loss is calculated between the labels and the corresponding input\_ids. All this is done by the following code:
\newline
$loss = loss\_fct(lm\_logits.view(-1, lm\_logits.size(-1)), \newline labels.view(-1))$

%% file: query_table.tex
\begin{table*}[h]
\begin{tabularx}{\textwidth}{X|X|X|X|X}
    \toprule
    {$Input Query$} &
    {$OneLook$} &
    {$Rev Dictionary$} &
    {$Want Words$} &
    {$WordAlchemy - T5$}
    \\
    \midrule
    ``To be unable to remember something" &
    forgets, forgets, bedwetter, hangover, misremember &
    blank, not having clue, lose, remind, illiterate &
    not remembering, confusion, forgets, blank, delusion  &
    forget, forgetfulness, amnesia, remembrance, forgetting
    \\
    ``to establish the identity of" &
    acertain, understand, assess, determine, recognize &
    settle, id, prove, set, disguise &
    reidentify, identity, anonymize, character, discover &
    identify, identified, recognize, distinguish, identity
    \\
    ``covered with water" &
    drowned, submerged, washed, drown, flood &
    deep, awash, swamp, wet, iced &
    marsh, afloat, soaked, flooding, water &
    flooded, swollen, drained, puddled, soaked
    \\
    ``workplace consisting of a room or building where movies are made" &
    studio, ropewalk, diary, door, gallery &
    hall, gallery, factory, theatre, labortory &
    location, cinema, auditorium, hall, studio &
    studio, theatre, cinema production, filmhouse, film station
    \\
\bottomrule
\end{tabularx}

\vspace{0.3em}
\caption{Comparison between queries of existing and T5 based Reverse Dictionaries }
\label{table:queries}

\end{table*}

%% file: results.tex
\section{Experimental Results and Analysis} 

\begin{figure}[ht]
\centering
\includegraphics[width=0.47\textwidth]{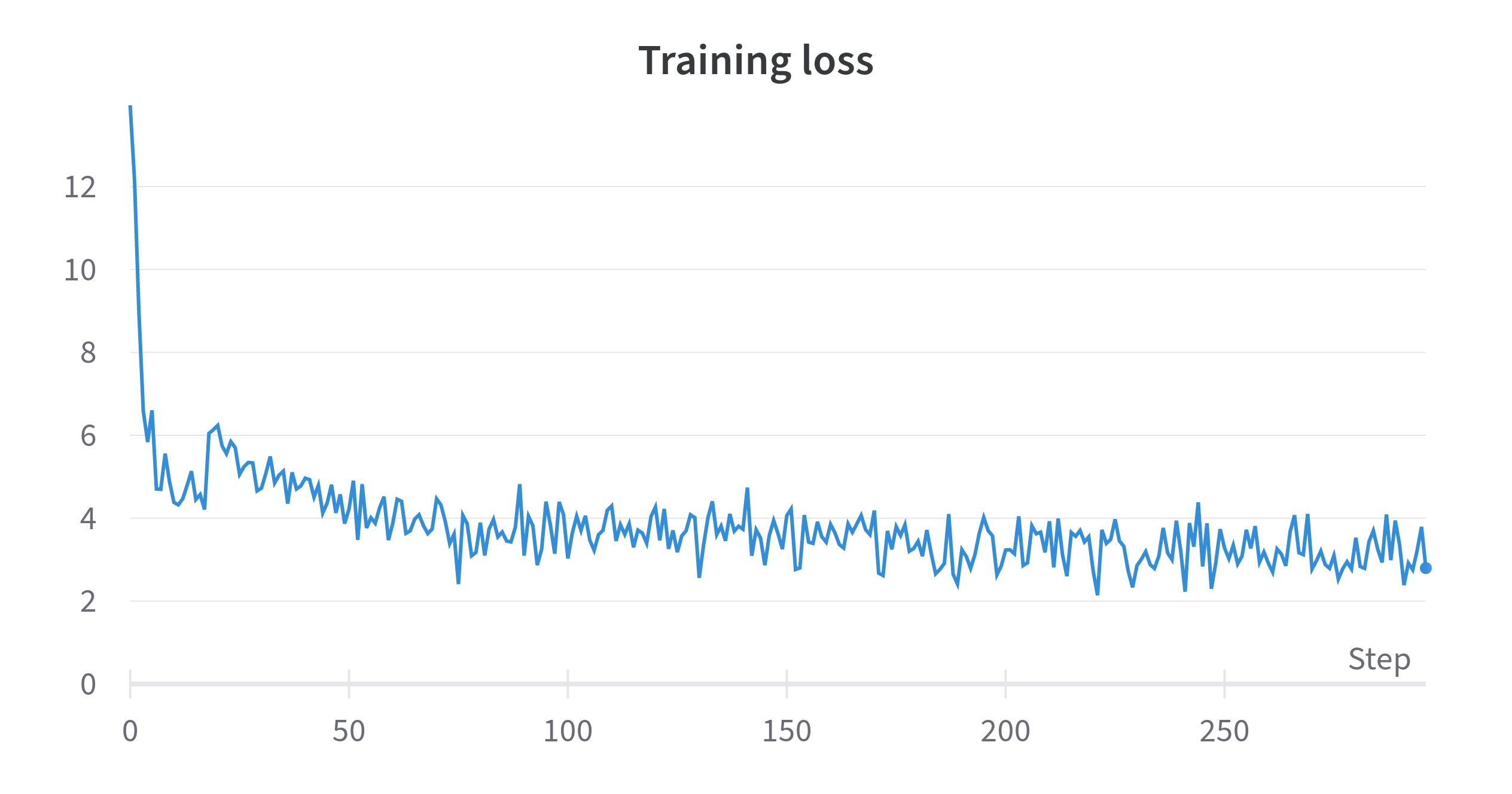}
\caption{mT5 training loss}
\end{figure}

\input{mono-lingual_table}

\begin{figure*}[h]
\centering
\includegraphics[width=1\textwidth]{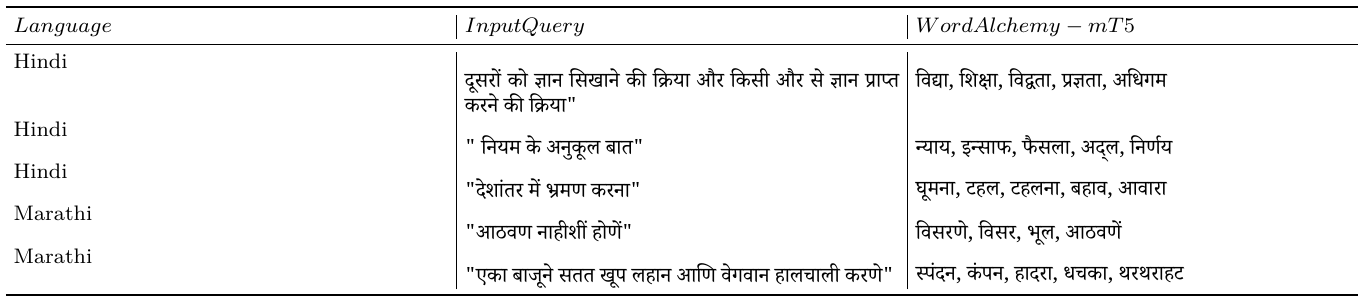}
\caption{Results on the Multi-lingual T5-based reverse dictionary.}
\label{table:multi-lingual-queries}
\end{figure*}

\subsection{Mono-lingual Reverse Dictionary}
Table \ref{table:mono-lingual} shows the findings and baseline techniques for monolingual reverse dictionary evaluation of T5. The most frequently employed reverse dictionary system, "OneLook," is indexed by over 1000+ dictionaries, including online dictionaries like Wikipedia and WordNet (Miller, 1995). As a result, the unseen definition test set's result is ignored. Whereas, results for "Mul-Channel" are referred from (Zhang et al., 2019). We can't exclude the word-definition pairings in the Definition set from OneLook and ReverseDictionary's databases because they have all of the WordNet definitions. As a result, they can only be evaluated based on the Description set as shown in Table 1. Models like BERT and RoBERTa show better results on the "seen" dataset. Despite its impressive performance on seen test sets, the Bert model fails to generalize to unseen test sets. On the other hand, we observe that T5 significantly outperforms these models. As observed from Table 1, T5 drastically improves performance over the description and unseen dataset as compared to Mul-Channel and OneLook dictionaries.

\subsection{Multi-lingual Reverse Dictionary}

Currently, there does not exist any multi-lingual reverse dictionary which has support for Indian Languages like Hindi and Marathi. We developed a multi-lingual reverse dictionary using mT5 transformer architecture.
Figure \ref{table:multi-lingual-queries} shows some queries and their outputs as produced by the model. Though it is difficult to achieve great results for Hindi and Marathi languages due to smaller data sets, we can see mT5 being pre-trained in such languages to overcome these data set challenges.

%% file: mono-lingual_table.tex
\begin{table}
\setlength{\tabcolsep}{0.7\tabcolsep}
\centering
\begin{tabularx}{0.5\textwidth}{p{2.3cm} | p{0.2cm}p{1.4cm} | p{0.2cm}p{1.4cm} | p{0.2cm}p{1.4cm}}
    \toprule
    {$Model$} &
    \multicolumn{2}{c|}{$Seen$} &
    \multicolumn{2}{c|}{$Unseen$} &
    \multicolumn{2}{c}{$Description$}
    \\
    \midrule
    OneLook &
    0 & .66/.94/.95 &
    - & - &
    5.5 & .33/.54/.76
    \\
    ReverseDictionary &
    4 & .30/.64/.80 &
    - & - &
    24 & .14/.60/.61
    \\
    Mul-Channel &
    16 & .20/.44/.71 &
    54 & .09/.29/.58 &
    2 & .32/.64/.88
    \\
    BERT &
    0 & .57/.86/.92 &
    18 & .20/.46/.64 &
    1 & .36/.77/.90
    \\
    RoBERTa &
    0 & .57/.84/.92 &
    37 & .10/.36/.60 &
    1 & .43/.84/.92
    \\
    T5 &
    - & .55/.88/.95 &
    - & .14/.58/.78 &
    - & .28/.88/.96
    \\
    \bottomrule
\end{tabularx}
\vspace{0.3em}
\caption{``Results on the English reverse dictionary
datasets. In each cell, the values are the “Median
Rank”, “Acc@1/10/100”." * results are from (Zhang et al., 2019). T5 achieves a significant performance boost}

\label{table:mono-lingual}

\end{table}

%% file: conclusion.tex
\section{Conclusion}

In this paper, we define the reverse dictionary task within the context of a text-to-text model framework and use the T5 model to predict the target word. Since T5 is an encoder-decoder model, it performs best across all downstream tasks as compared to the decoder only an encoder only transformer models. It also being pre-trained on the large C4 dataset, helps to correctly predict the target word even if it was not present in the training dataset. Human-written descriptions are sometimes long and not so on-point accurate. This problem has been addressed by using a transformer model which uses self-attention to correctly depict the meaning of the inputs.